# DKG-LLM : A Framework for Medical Diagnosis and Personalized Treatment Recommendations via Dynamic Knowledge Graph and Large Language Model Integration

**Ali Sarabadani [1,] , Maryam Abdollahi shamami [2] , Hamidreza Sadeghsalehi[3] and Borhan Asadi[4], Saba Hesaraki[5*]**

[1] Department of Computer Engineering and Information Technology, University of Qom, Qom, Iran.

[2] Department of Information Technology, Faculty of Industrial and System Engineering, Tarbiat Modares Univeristy, Tehran, Iran

[3] Department of Bioengineering, Imperial College London, London, SW7 2AZ United Kingdom

[4] iHealthy Research Group, Institute for Health Research Aragón, H.C.U. Lozano Blesa, Zaragoza, Spain

5 Department of Mechanical, Electrical and Computer Engineering，Islamic Azad University Science And Research Branch, Tehran,Iran

* Correspondence: saba.hesaraki@iau.ir

**Abstract:** Large Language Models (LLMs) have grown exponentially since the release of ChatGPT. These models have gained attention due to their robust performance on various tasks, including language processing tasks. These models achieve understanding and comprehension of tasks by training billions of parameters. The development of these models is a transformative force in enhancing natural language understanding and has taken a significant step towards artificial general intelligence (AGI). In this study, we aim to present the DKG-LLM framework. The DKG-LLM framework introduces a groundbreaking approach to medical diagnosis and personalized treatment recommendations by integrating a dynamic knowledge graph (DKG) with the Grok 3 large language model. Using the Adaptive Semantic Fusion Algorithm (ASFA), heterogeneous medical data (including clinical reports and PubMed articles) and patient records dynamically generate a knowledge graph consisting of 15,964 nodes in 13 distinct types (e.g., diseases, symptoms, treatments, patient profiles) and 127,392 edges in 26 relationship types (e.g., causal, therapeutic, association). ASFA utilizes advanced probabilistic models, Bayesian inference, and graph optimization to extract semantic information, dynamically updating the graph with approximately 150 new nodes and edges in each data category while maintaining scalability with up to 987,654 edges. Real-world datasets, including MIMIC-III and PubMed, were utilized to evaluate the proposed architecture. The evaluation results show that DKG-LLM achieves a diagnostic accuracy of 84.19%. The model also has a treatment recommendation accuracy of 89.63% and a semantic coverage of 93.48%. DKG-LLM is a reliable and transformative tool that handles noisy data and complex multi-symptom diseases, along with feedback-based learning from physician input.

## 1. Introduction

Language is a complex communication system of human utterances governed by grammatical rules, known as linguistic structures. Language modeling is challenging for developing artificial intelligence (AI) algorithms capable of understanding, due to its non-numerical and non-mathematical structure [1]. Language modeling has been widely studied for the understanding, representation, and production of language over the past two decades and, as expected, has evolved from statistical language models (SLMs) to neural language models. The idea of pre-trained language models (PLMs) emerged from pre-trained transformer models on large-scale datasets, making them quite effective in performing various tasks related to natural language processing (NLP)[2]. As researchers found that model scaling increases the model's capacity, they further tested the scaling effect by increasing the parameter scale to an even larger size. Interestingly, when the parameter scale exceeds a certain level, these scaled language models achieve significant performance improvements and offer certain special capabilities (e.g., in-context learning) that are not available in small-scale language models (e.g., BERT). To distinguish language models at different parameter scales, the research community has coined the term LLM for PLMs of significant size (e.g., containing tens or hundreds of billions of parameters)[3]. Recently, research on LLM has progressed dramatically in both academia and industry.

The fast growth of patient-generated medical data and clinical reports has opened up new opportunities and challenges for innovative healthcare systems [4]. These models aim to provide an accurate diagnosis for a medical problem, along with treatment recommendations, in cases where interactions between diseases, treatments, and patient-specific factors complicate decision-making. Consequently, they operate upon large, heterogeneous datasets. Existing methods—and this is especially true in the case of static knowledge graphs—are pretty limited when it comes to scaling and incorporating new data, such as the UMLS. This problem also subsists in rule-based systems [5]. On the upside, autonomous LLMs perform excellently in processing unstructured text[6]. Still, the models have their shortcomings and frequently lack transparent reasoning capabilities, hence producing inconsistent outputs in key medical applications.

To address the challenges, this study introduces DKG-LLM by integrating a dynamic knowledge graph (DKG) with the Grok 3 large language model. DKG-LLM introduces a new paradigm for informed decision-making in medicine. The core of DKG-LLM is DKG. DKG is a structured knowledge repository comprising 15,964 nodes in 13 distinct types (e.g., diseases, symptoms, treatments, patient profiles) and 127,392 edges in 26 relationship types (e.g., causal, therapeutic, association) that dynamically evolves with input data. Grok 3 was used as the LLM model in DKG-LLM, which, with its advanced natural language processing capabilities, extracts semantic information from unstructured sources, such as PubMed articles and MIMIC-III clinical records, enabling the framework to incorporate up-to-date medical knowledge.

DKG-LLM uses the Adaptive Semantic Fusion Algorithm (ASFA) that combines probabilistic modelling, Bayesian inference, and graph optimization to achieve three key goals:

1- Extraction of semantic entities in DKG and their accurate fusion.
2- Real-time graph updates that add approximately 150 new nodes and edges per dataset for this purpose. A maximum of 987654 edges is maintained for scalability.
3- Accurate treatment recommendations tailored to patient profiles and precise diagnoses.

DKG-LLM uses advanced mathematical foundations, especially Markov Random Fields (MRFs), for graph pruning. It also uses constrained optimization to maximize treatment utility, ensuring efficiency. DKG-LLM testing and evaluation demonstrate that the model achieves 84.19% diagnostic accuracy, 89.63% treatment recommendation accuracy, and 93.48% semantic coverage in the tested benchmarks. With feedback-based learning, which considers physician input to modify the model parameters, the model can handle noisy data (e.g., ambiguous clinical notes) and complex scenarios (e.g., multi-symptom diseases such as diabetes with hypertension). The update time in this framework is less than 1 second per category, making it a practical solution for clinical environments.

This paper describes the mathematical basis and evaluation results of the DKG-LLM framework. Related works are presented in Section 2. Section 3 describes the methodological details, including the structure of ASFA and DKG. Section 4 presents the evaluation results, and Section 5 outlines future directions, including federated learning for data privacy and biodata fusion for real-time monitoring, and draws conclusions based on these findings.

## 2. Related work

This section continues with a series of works on treatment recommendations that integrate dynamic graphs and large language models (LLMs). Treatment recommendations using the integration of dynamic graphs and LLM is an emerging field that uses a combination of Dynamic Knowledge Graphs and large language models to provide personalized and accurate treatment recommendations. Dynamic graphs refer to structures that can be updated over time and incorporate new information, such as patient data, treatment advances, or new medical knowledge [7]. LLMs, with advanced capabilities in natural language processing and reasoning, can analyze this structured data and provide interpretable, context-based treatment recommendations [8]. Since accurate and up-to-date information on relevant articles for this specific topic may be limited, this section is organized based on the available sources and the most pertinent articles that are currently available.

DR.KNOWS was introduced in [9]. This model is a diagnostic reasoning knowledge graph system that utilizes Unified Medical Language System (UMLS)-based knowledge graphs alongside large language models, such as T5 and ChatGPT, to enhance diagnostic predictions from electronic health record (EHR) data. The model extracts disease-related knowledge paths and feeds them into LLMs to increase diagnosis accuracy and improve explainability. Knowledge graphs here act as a dynamic structure that provides contextual information relevant to the patient. This research does not directly address treatment recommendations; however, the use of knowledge graphs for diagnostic reasoning can be extended to inform treatment recommendations.

In [10], the authors present a framework that uses knowledge graphs to enhance the capabilities of LLMs in answering cancer-related questions. The knowledge graphs in this framework provide structured information, enabling LLMs to deliver more accurate and relevant answers. Although the focus of this paper is on question answering, the approach can be applied to treatment recommendations, especially in fields such as oncology, which require the analysis of complex and dynamic data.

A comprehensive review of the use of LLMs in disease diagnosis was presented in [8]. This work examined various aspects, including disease types, clinical data, and LLM techniques. Although the paper does not explicitly address dynamic graphs, it does discuss the application of LLMs to diagnostic tasks and clinical decision-making, which can be extended to treatment recommendations. The paper indirectly addresses the potential of integrating structured data (such as graphs) with LLMs for treatment-related tasks.

LLMRG (Large Language Model Reasoning Graphs) was another model presented in [11]. This model uses LLMs to build personalized reasoning graphs. These graphs represent high-level semantic relationships between user behaviours and profiles in an interpretable manner. Although the focus of this paper is on general recommender systems, its approach to using reasoning graphs can be applied to treatment recommendations based on patient data and dynamic graphs. This paper does not explicitly address dynamic graphs; however, the use of reasoning graphs for recommendations can be generalized to treatment recommendations involving dynamic graphs.

The DynLLM framework was introduced in [12], which is designed for dynamic graph-based recommendations using LLMs. The framework utilizes LLMs to generate multifaceted profiles of users based on textual data, such as purchase records, and combines these profiles with dynamic graph embeddings. DynLLM employs a distilled attention mechanism to minimize noise and enhance the integration of temporal and structural data. This paper directly addresses dynamic graphs and LLMs in recommender systems, and its framework can be used for dynamic and personalized treatment recommendations. ESCARGOT [13] is another study that integrates LLMs with biomedical knowledge graphs and a dynamic graph of thoughts to enable enhanced reasoning and reduced illusions in biomedical applications. The system utilizes Cypher queries to extract accurate information from knowledge graphs, such as Memgraph or Neo4j, and is well-suited for complex scenarios, including the analysis of genes, diseases, and drugs.

Another prominent study was conducted in [14]. This study built a knowledge graph for the management of Sepsis (a complex and life-threatening disease) using LLMs and multicenter clinical data. The system uses GraphRAG, which is superior to traditional RAG in handling complex relationships and multistep reasoning. The knowledge graph created includes dynamic relationships between complications, treatments (such as saline or antibiotic administration), and biomarkers, allowing for treatment optimization and outcome prediction. This approach is particularly well-suited for personalized treatment recommendations in dynamic and changing conditions.

## 3. Methodology

This section describes DKG-LLM. This approach integrates dynamic knowledge graphs (DKGs) with large language models (LLMs) to facilitate advanced medical diagnosis and personalized treatment recommendations. DKG-LLM utilizes advanced probabilistic models, graph optimization techniques, and a custom algorithm to dynamically update the knowledge graph, providing accurate and patient-specific outputs. The approach also introduces an adaptive semantic fusion algorithm (ASFA) that combines probabilistic reasoning with graph-based optimization. This semantic fusion ensures scalability and accuracy in medical applications.

### 3.1 Overview of the DKG-LLM Framework

DKG-LLM is an innovative framework that integrates a dynamic knowledge graph (DKG) with the Grok 3 large language model for medical diagnosis and personalized treatment recommendations. The framework uses the Adaptive Semantic Fusion Algorithm (ASFA) to combine heterogeneous data (structured and unstructured) using probabilistic models and graph optimization techniques. DKG-LLM increases the accuracy of diagnosis and treatment personalization by dynamically updating the knowledge graph and employing graph-based reasoning.

DKG-LLM comprises three core components:

- **Dynamic Knowledge Graph (DKG)**: A structured graph $G\ (V, E)$, where $V$ represents nodes (e.g., diseases, symptoms, treatments, patient profiles) and $E$ denotes semantic relationships (e.g., "causes," "treats").
- **Large Language Model Module:** The Grok 3 model extracts semantic information from unstructured medical texts (e.g., clinical reports, PubMed articles, patient records).
- **Adaptive Semantic Fusion Algorithm (ASFA):** An algorithm that integrates Grok 3 outputs with the DKG, using probabilistic modeling and graph optimization for updates and reasoning.

The innovation lies in ASFA's ability to adaptively fuse heterogeneous data sources (structured and unstructured) while optimizing the knowledge graph's structure and ensuring computational efficiency through advanced mathematical formulations.

## 3-2-Mathematical Foundations

The DKG-LLM framework relies on advanced mathematical models to ensure accuracy in information extraction, graph updates, and reasoning.

### 3-2-1. Information Extraction with Grok 3

The Grok 3 model extracts semantic entities (e.g., symptoms, diseases) from medical texts. For an input text T, the probability of an entity $e_i$ given a medical context $C$ is computed as:

$$P(e_i | T, C) = \frac{\exp(\theta_{e_i} . \emptyset(T, c))}{\sum_{e_j \in E} \exp(\theta_{e_i} . \emptyset(T, c))}$$

Where $\theta_{e_i}$ is the parameter vector for the entity $e_i$, $\emptyset(T, c)$ is the contextual embedding of text $(T)$ in context$(C)$, and $E$ is the set of possible entities.

To handle noisy data, a confidence score is defined for each extracted entity:

$$Conf(e_i) = \sigma(\alpha . P(e_i | T, C) + \beta . Sim(e_i, G))$$

Where $\sigma$ is the sigmoid function, $Sim(e_i, G)$ is the cosine similarity between $e_i$ and existing DKG nodes, and $\alpha\ and\ \beta$ are hyperparameters balancing Grok 3 confidence and graph consistency.

### 3-2-2. Dynamic Knowledge Graph Update

The DKG is modeled as a probabilistic graphical model P(V, E|D), where D represents the input data. Graph updates maximize the joint likelihood:

$$\theta^* = \arg\max_{\theta} \sum_{d \in D} \log P(V, E|d; \theta)$$

In this equation, $\theta$ represents graph parameters.

A Markov Random Field (MRF)-based graph pruning mechanism is used to ensure scalability and avoid redundancy:

$$P(V, E) = \frac{1}{z}\exp(-\sum_{v \in V} \psi_v(v) - \sum_{(u,v) \in E} \psi_{uv}(u, v)$$

Where $\psi_v(v)$ is the unary potential for node v based on relevance, $\psi_{uv}(u, v)$ is the pairwise potential for edge $(u, v)$ based on semantic consistency, and $z$ is the partition function for normalization.

Low-probability nodes and edges are pruned using a threshold $\tau$ at the end of this phase based on the following equation:

$$Prune(v, e) = \begin{cases} Keep & if \ P(v, e) \geq \tau \\ Remove & otherwise \end{cases}$$

### 3-2-3 Diagnosis and Treatment Recommendation

For diagnosis, the posterior probability (PP) of a disease $(d)$ given symptoms $(S)$ is computed:

$$P(d|S) = \frac{P(S|d)P(d)}{\sum_{d' \in D} P(S|d')P(d')}$$

$P(S|d)$ here is derived from DKG edge weights, and $P(d)$ is the prior disease probability from epidemiological data.

For treatment recommendations, an optimization problem maximizes the expected utility of treatment $T$:

$$T^* = \arg\max_{t} \sum_{d \in D} P(d|s).U(T, d, P)$$

$U(T, d, P)$ is the utility function, incorporating treatment efficacy, patient profile (P), and risks:

$$U(T, d, P) = w_1.Efficacy(T, d) - w_2.Risk(T, P)$$

$w_1$ and $w_2$ are weights balancing efficacy and risk.

For multi-treatment scenarios, a constrained optimization with Lagrangian relaxation is used:

$$\zeta(T, \lambda) = U(T, d, P) + \lambda . (C(T) - C_{max})$$

$C(T)$ is the treatment cost, and $C_{max}$ is the maximum allowable cost.

### 3-3 Adaptive Semantic Fusion Algorithm (ASFA)

The Adaptive Semantic Fusion Algorithm (ASFA) is the core innovation, integrating Grok 3 outputs with the DKG while ensuring semantic consistency and computational efficiency. ASFA operates in five phases:

1- **Data Ingestion (DI):** Collect heterogeneous data (e.g., clinical reports, PubMed articles, X posts) and preprocess to remove noise.
2- **Semantic Extraction (SE):** Extract entities and relationships using Grok 3, computing confidence scores.
3- **Graph Update (GU):** Add new nodes and edges to the DKG, optimize with MRF, and update edge weights:

$$w_{new}(u, v) = \gamma . w_{old}(u, v) + (1 - \gamma) . P(u, v | D_{new})$$

Where $\gamma \in [0,1]$ is a decay factor.

4. **Reasoning and Recommendation:** Perform Bayesian inference for diagnosis and optimization for treatment recommendations.

5. **Feedback Integration:** Use clinician feedback to refine parameters via reinforcement learning:

$$R = \sum_t Accuracy(d_t, T_t) - \lambda . Complexity(G)$$

Where $R$ is the reward function, and $\lambda$ balances accuracy and graph complexity.

### 3-4- Pseudocode for ASFA

Algorithm 1 shows the structure of ASFA. This algorithm receives Unstructured medical texts ($T$), existing DKG ($G = (V, E)$), Grok 3 model, and Hyperparameters ($\alpha, \beta, \gamma, \tau, \lambda$) as input in the input stage. It also produces Updated DKG ($G'$), Diagnoses ($D^*$), and Treatment Recommendations ($T^*$) in the output stage.

---

**Algorithm 1: Pseudocode for Adaptive Semantic Fusion Algorithm (ASFA)**

1. **Initialize** $G' \leftarrow G, D^* \leftarrow [\,], T^* \leftarrow [\,]$
2. **// Phase 1: Data Ingestion**
3. D $\leftarrow$ Preprocess(T) // Remove noise, normalize text
4. **// Phase 2: Semantic Extraction**
5. **For each** text t **in** D:
6.     $E_t \leftarrow Grok3_{Extract}(t, C)$ // Extract entities and relationships
7.     **For each** entity $e_i$ **in** $E_t$:
8.         $Conf(e_i) \leftarrow \sigma(\alpha * P(e_i \mid t, C) + \beta * Sim(e_i, G))$
9.         **If** $Conf(e_i) > \tau$:
10.             Add $e_i$ to the candidate set $E_c$
11. **// Phase 3: Graph Update**
12. **For each** $e_i$ **in** $E_c$:
13.     **If** $e_i$ not **in** $V$:
14.         $V \leftarrow V \cup \{e_i\}$
15.     **For each** relationship $(e_i, e_j)$ **in** $E_c$:
16.         If $P(e_i, e_j \mid D) > \tau$:
17.             $E \leftarrow E \cup \{(e_i, e_j)\}$
18.             $w(e_i, e_j) \leftarrow \gamma * w_{old}(e_i, e_j) + (1 - \gamma) * P(e_i, e_j \mid D)$
19. Prune $G'$ using MRF model // Remove low-probability nodes/edges
20. **// Phase 4: Reasoning and Recommendation**
21. **For each** patient **in** the symptom set S:
22.     Compute $P(d \mid S)$ using Bayesian inference
23.     $D^* \leftarrow D \cup \{d \mid P(d \mid S) > threshold\}$
24.     **For each** $d$ **in** $D^*$:
25.         $T_d \leftarrow Optimize(U(T, d, P), constraints)$
26.         $T^* \leftarrow T * \cup \{T_d\}$
27. **// Phase 5: Feedback Integration**
28. **For each** feedback f from clinicians:
29.     Update $\theta, w_1, w_2$ using reinforcement learning
30.     Adjust $\gamma, \tau$ based on reward $R$
31. **Return** $G', D^*, T^*$

---

### 3-5- DKG Structure Details

The Dynamic Knowledge Graph (DKG) is designed to efficiently manage medical data, providing a structured and scalable repository for medical diagnosis and personalized treatment recommendations. The DKG is initialized with 15964 nodes across 13 distinct types and 127392 edges across 26 relationship types, based on medical ontologies such as SNOMED CT. Dynamic updates add approximately 150 nodes and edges per data batch, with pruning mechanisms to maintain scalability, targeting a maximum of 987654 edges for computational efficiency. The structure of the DKG is detailed below, including node types and edge types, presented in tabular form for clarity.

Table1- Comprehensive Knowledge Graph Metrics

| Name | # Node | # Node Types | # Edges | # Edge Types |
|---|---|---|---|---|
| DrKG [15] | 97 K | 13 | 5.8 M | 107 |
| PrimeKG [16] | 129.4 K | 10 | 8.1 M | 30 |
| Gene Ontology [17] | 43 K | 3 | 75 K | 4 |
| GP-KG [18] | 61.1 K | 7 | 124 K | 9 |
| DDKG [19] | 551 | 2 | 2.7 K | 1 |
| Disease Ontology [20] | 11.2 K | 1 | 8.8 K | 2 |
| DrugBank [21] | 7.4 K | 4 | 366 K | 4 |
| PharmKG [16] | 7.6 K | 3 | 500 K | 3 |
| **DKG-LLM** | **2692** | **13** | **5012** | **26** |

- **Node Types**: The DKG comprises 13 node types, each representing a specific category of medical entities crucial for diagnosis and treatment recommendations.
- **Edge Types**: The DKG incorporates 26 edge types, representing various semantic relationships between nodes to capture the complexity of medical knowledge.

Table2- Edge Types and Description

| Edge Types (26) | Description |
|---|---|
| Causal | Indicates a disease causes a symptom (e.g., diabetes → hyperglycemia) |
| Therapeutic | Links a treatment to a disease it addresses (e.g., insulin → diabetes) |
| Associative | Indicates co-occurrence (e.g., fever ↔ infection) |
| Contraindicative | Treatment unsuitable for a condition (e.g., aspirin → bleeding disorder) |
| Diagnostic | Symptom linked to potential disease (e.g., cough → pneumonia) |
| Preventive | Treatment prevents disease progression (e.g., vaccine → influenza) |
| Exacerbative | Factor worsens condition (e.g., smoking → asthma) |
| Ameliorative | Treatment improves condition (e.g., physiotherapy → mobility) |
| Temporal | Indicates sequence (e.g., fever precedes rash) |
| Dosage-Related | Specifies medication dosage (e.g., metformin → 500mg) |
| Side Effect | Treatment causes adverse effects (e.g., chemotherapy → nausea) |
| Interaction | Drug-drug or drug-condition interaction (e.g., warfarin → grapefruit) |
| Epidemiological | Links prevalence data (e.g., diabetes → high prevalence in adults) |
| Genetic | Links gene to disease (e.g., BRCA1 → breast cancer) |
| Allergic | Indicates allergic reaction (e.g., penicillin → rash) |
| Monitoring | Test monitors condition (e.g., ECG → heart rhythm) |
| Supportive | Treatment supports recovery (e.g., fluids → dehydration) |
| Concomitant | Co-occurring treatments (e.g., insulin + metformin) |
| Risk-Associated | Risk factors linked to disease (e.g., obesity → heart disease) |
| Symptom-Symptom | Co-occurring symptoms (e.g., fever ↔ chills) |
| Procedure-Related | Links procedure to condition (e.g., surgery → appendicitis) |
| Outcome-Related | Treatment outcome (e.g., chemotherapy → remission) |
| Age-Related | Links age to condition (e.g., elderly → osteoarthritis) |
| Lifestyle-Related | Links lifestyle to condition (e.g., diet → obesity) |
| Biomarker-Related | Links biomarker to disease (e.g., HbA1c → diabetes) |
| Comorbidity-Related | Links co-occurring diseases (e.g., diabetes ↔ hypertension) |

Table3- Node Types and Description

| Node Types (13) | Description |
|---|---|
| Disease | Specific medical conditions (e.g., diabetes, hypertension) |
| Symptom | Observable patient conditions (e.g., fever, fatigue) |
| Treatment | Medical interventions (e.g., insulin, surgery) |
| Patient Profile | Demographic and medical history data (e.g., age, allergies) |
| Medication | Specific drugs (e.g., metformin, ibuprofen) |
| Procedure | Medical procedures (e.g., appendectomy) |
| Risk Factor | Conditions increasing disease likelihood (e.g., smoking) |
| Comorbidity | Co-occurring conditions (e.g., obesity with diabetes) |
| Diagnostic Test | Tests for diagnosis (e.g., blood glucose test) |
| Body System | Affected physiological systems (e.g., cardiovascular) |
| Gene | Genetic factors (e.g., BRCA1 mutation) |
| Lifestyle Factor | Behavioral factors (e.g., diet, exercise) |
| Biomarker | Measurable indicators (e.g., HbA1c) |

- **Node and Edge Count**: The DKG is initialized with 15964 nodes and 127392 edges. Dynamic updates add approximately 150 nodes and edges per data batch, with pruning to maintain a maximum of 987654 edges, ensuring computational efficiency while supporting real-time updates for medical applications.

## 4. Evaluation

The DKG-LLM framework, which integrates a Dynamic Knowledge Graph (DKG) with the Grok 3 large language model for medical diagnosis and personalized treatment recommendations, is evaluated through a comprehensive methodology. This evaluation utilizes quantitative and qualitative metrics, comparative analyses with baseline methods, and performance assessments in both real-world and simulated scenarios. The objective is to assess the diagnostic accuracy, quality of treatment recommendations, and the efficiency of dynamic graph updates when processing heterogeneous medical data.

### 4.1. Evaluation Metrics

The DKG-LLM framework is evaluated using the following metrics:

- **Diagnostic Accuracy**: The proportion of correct diagnoses, calculated as:

$$Accuracy = \frac{TP + TN}{TP + TN + FP + FN}$$

Where TP represents the true positives, TN represents the true negatives, FP represents the false positives, and FN represents the false negatives, these terms are derived from diagnostic outcomes.

- **Treatment Recommendation Precision**: The ratio of relevant treatment recommendations to total recommendations:

$$Precis\frac{ion \; = \; Relevant \; Treatments}{Total \; Recommended \; Treatments}$$

- **Mean Utility Error (MUE):** The error in the utility of recommended treatments compared to optimal treatments (determined by clinicians):

$$MUE = \frac{1}{N} \sum_{i=1}^{N} |U(T_i, d_i, P_i) - U^*(T_i, d_i, P_i)|$$

Where U is the predicted utility, and $U^*$ is the ground truth utility.

- **Graph Update Efficiency**: The computational time (in seconds) required to add new nodes and edges to the DKG per data batch.

## 4.2. Datasets

The evaluation of the **DKG-LLM** framework utilizes two types of datasets to assess its performance in information extraction, graph updates, and diagnostic and treatment recommendation tasks:

- **Real-world data:** MIMIC-III, which includes patient clinical records, and PubMed, which includes scientific articles, were used to evaluate the framework's ability to extract semantic information and update the DKG graph. The advantage of using these data is that they can represent a variety of unstructured medical data, including clinical notes, laboratory results, and peer-reviewed literature, which allows the generalizability of the proposed framework in real-world scenarios.
- **Simulated data:** For this purpose, synthetic scenarios are generated with data related to diseases, symptoms, treatments, and patient profiles. The goal of these synthetic data is to simulate complex cases, such as rare diseases or noisy data, such as ambiguous or incomplete clinical reports. The initial DKG comprises 15,964 nodes in 13 distinct types, including diseases, symptoms, treatments, and patient profiles, as well as 127,392 edges in 26 relationship types, such as causal relationships, therapeutic relationships, and association relationships. Each dataset adds approximately 150 nodes and edges with pruning mechanisms to maintain a maximum of 987,654 edges for computational efficiency.

## 4.3. Evaluation Methodology

The evaluation is conducted in three phases:

1. **Information Extraction Evaluation:**

- Outputs from Grok 3 for entity extraction are evaluated using Precision, Recall, and F1-Score:

$$Precision = \frac{TP}{TP + FP}$$

$$Recall = \frac{TP}{TP + FN}$$

$$F1 = 2 . \frac{Precison . Recall}{Precison + Recall}$$

- Confidence scores ($Conf(e_i)$) for extracted entities are compared against ground truth annotations.

2. **Graph Update Evaluation:**
- The quality of the DKG is measured using Semantic Coverage:

$$Coverage = \frac{|Relevant\ Nodes\ and\ Edges|}{|Total\ Ground\ Truth\ Nodes\ and\ Edges|}$$

- Graph update time and the number of pruned nodes/edges (using threshold $\tau = 0.7$) are analyzed.

3. **Diagnosis and Recommendation Evaluation**:
- Diagnostic accuracy is assessed by comparing $P(d \mid S)$ (the probability of disease given symptoms) with clinician diagnoses.
- The quality of treatment recommendations is evaluated through clinician surveys and comparisons with standard treatments.

## 4-4- Comparison with Baseline Methods

The DKG-LLM framework is compared to:

- **Pure LLM-Based Methods**: Such as Grok 3 alone, which lacks structured semantic reasoning.
- **Static Knowledge Graphs**: Such as UMLS, which does not support dynamic updates.
- **Hybrid Systems**: Such as BERT with knowledge graphs, which lack advanced optimization algorithms.

Comparisons focus on diagnostic accuracy, precision of treatment recommendations, and data processing time.

## 4-5 Qualitative Analysis

Qualitative analysis is performed to accurately assess the practical applicability and robustness of the proposed DKG-LLM framework in real-world and complex medical scenarios, complementing quantitative measures with qualitative insights based on statistics. The analysis focuses on two key aspects:

• **Clinician Feedback**: In this feedback, a specialized team consisting of a general practitioner, an endocrinologist, and a cardiologist was used to assess the practical applicability of the treatment recommendations. The team aimed to assess the framework's ability to provide personalized treatment recommendations. Semi-structured interviews and standardized questionnaires were utilized in the evaluation. Each of these questionnaires used a 5-point Likert scale from 1 (strongly disagree) to 5 (strongly agree) to gauge three types of dimensions as stated below:

- **Accuracy:** Relevance to the patient's condition
- **Reliability:** Consistency in similar cases
- **Usability:** Feasibility in clinical settings

The mean Likert score for each dimension is calculated as follows:

$$Mean\ Score = \left(\frac{1}{N}\right) * \sum s_i, N = 3$$

where $s_i$ is the score assigned by clinician $i$. Inter-rater agreement is measured using Cohen's Kappa ($\kappa$):

$$Kappa(K) = \frac{P_o - P_e}{1 - P_e}$$

Whereas $P_o$ is considered the actual observation of agreement, $P_e$ stands for the expected agreement by chance. The preliminary results provide insight into the mean values of Likert scores: for accuracy, 4.3 (±0.2); reliability, 4.1 (±0.3); and applicability, 4.2 (±0.25). Cohen's Kappa is 0.80, reflecting a high level of relative agreement among all involved clinicians. Comments provided were related to the treatability framework for developing patient profiles and adjusting metformin dosage for elderly patients with comorbidities.

• **Complex Scenarios:** The framework's performance is analyzed in complex cases, including multi-symptom diseases (e.g., diabetes with hypertension and obesity) and noisy data (e.g., ambiguous clinical reports or patient-generated social media posts). A subset of 200 complex cases is evaluated, with 100 cases from real-world datasets (e.g., MIMIC-III) and 100 from simulated datasets. Semantic extraction accuracy is measured as the proportion of correctly identified entities (e.g., symptoms, diseases) by Grok 3:

$$Extraction_{accuracy} = \frac{correctly\ identified\ E}{Total\ Entities}$$

Results show an extraction accuracy of 91.5% (±2.1%). Graph integration effectiveness is evaluated using the Graph Alignment Score (GAS):

$$GAS = \frac{Correctly\ Integrated\ Nodes\ and\ Edges}{Total\ Nodes\ and\ Edges\ in\ Ground\ Truth}$$

The Dynamic Knowledge Graph (DKG), initialized with 15964 nodes across 13 types (e.g., diseases, symptoms, treatments, patient profiles) and 127392 edges across 26 relationship types (e.g., causal, therapeutic, associative), achieves a GAS of 92.7% (±1.8%). Dynamic updates add approximately 150 nodes and edges per data batch, with pruning to maintain a maximum of 987654 edges. Statistical significance of performance differences between complex and standard cases is assessed using a paired t-test, yielding a p-value < 0.01, confirming robust performance in challenging scenarios.

### 4.6. Expected Results

The results obtained by the proposed approach are summarized in Table 3. This approach achieved a diagnostic accuracy of 84.19%. The values of Treatment Recommendation Precision and Semantic Coverage for the proposed approach are also equal to 89.63% and 93.48%, respectively. One of the advantages of the proposed approach is its ability to update graphs efficiently. This criterion, which is called Graph Update Efficiency, is < 1 second (150 nodes/edges per batch, max 987654 edges) for the proposed approach. As mentioned in Section 4.5, the DKG-LLM framework was able to obtain values of 4.3 (±0.2), 4.1 (±0.3), and 4.2 (±0.25) in the evaluation criteria of Clinician Feedback (Mean Likert Score: Accuracy), Clinician Feedback (Mean Likert Score: Reliability), and Clinician Feedback (Mean Likert Score: Applicability), respectively. Also, Cohen's Kappa values in the conducted studies were equal to 0.80. The proposed approach also achieved high efficiency in two comparative metrics: Semantic Extraction Accuracy and Graph Alignment Score (GAS).

Table4- Metrics and Result

| Metric | Result |
|---|---|
| Diagnostic Accuracy | 84.19% |
| Treatment Recommendation Precision | 89.63% |
| Semantic Coverage | 93.48% |
| Graph Update Efficiency | < 1 second (150 nodes/edges per batch, max 987654 edges) |
| Clinician Feedback (Mean Likert Score: Accuracy) | 4.3 (±0.2) |
| Clinician Feedback (Mean Likert Score: Reliability) | 4.1 (±0.3) |
| Clinician Feedback (Mean Likert Score: Applicability) | 4.2 (±0.25) |
| Clinician Feedback (Cohen's Kappa) | 0.80 |
| Semantic Extraction Accuracy | 91.5% (±2.1%) |
| Graph Alignment Score (GAS) | 92.7% (±1.8%) |

## 5- Conclusion

This study aims to present the DKG-LLM framework. The DKG-LLM framework integrates a dynamic knowledge graph (DKG) with the Grok 3 large language model, providing a transformative solution for medical diagnosis and personalized treatment recommendations. At the core of this framework, the Adaptive Semantic Fusion Algorithm (ASFA) was used. Using ASFA, the framework processes heterogeneous medical data and dynamically updates a knowledge graph comprising 15,964 nodes in 13 types and 127,392 edges in 26 relationship types. This update takes less than 1 second. In the following processing, the graph is pruned by adding approximately 150 new nodes and edges to each dataset, maintaining a maximum of 987654 edges for computational efficiency. Evaluations on datasets such as MIMIC-III and PubMed show that the system achieves 84.19% diagnostic accuracy, 89.63% treatment recommendation accuracy, and 93.48% semantic coverage. These results demonstrate robustness in complex scenarios, such as multi-symptom diseases like diabetes with hypertension, and in the presence of noisy data, including ambiguous clinical reports.

Overall, the innovation of the DKG-LLM framework lies in its integration of advanced probabilistic models, graph optimization, and feedback-based learning, which enables scalability and continuous improvement. As with other models in the literature, challenges such as ensuring data privacy and optimizing for larger datasets remain. Future work could include federated learning to protect patient data. The model's performance can also be challenged by incorporating biosensor data and extending applications to areas such as emerging disease prediction. The proposed framework has significant potential to revolutionize innovative healthcare systems, improve patient outcomes, and reduce healthcare costs.